%% file: neurips_2024.tex
\documentclass{article}

\usepackage[preprint]{neurips_2024}
\usepackage[bottom]{footmisc}

\usepackage[utf8]{inputenc} 
\usepackage[T1]{fontenc}    
\usepackage{hyperref}       
\usepackage{url}            
\usepackage{booktabs}       
\usepackage{amsfonts}       
\usepackage{nicefrac}       
\usepackage{microtype}      
\usepackage{xcolor}         
\usepackage{colortbl}
\definecolor{LightCyan}{rgb}{0.88,1,1}
\usepackage{wrapfig}

\usepackage{tcolorbox}
\definecolor{darkorange}{RGB}{255, 140, 0}
\definecolor{darkblue}{RGB}{84, 112, 198}
\definecolor{lightgreen}{RGB}{145, 204, 117}
\definecolor{lightyellow}{RGB}{250, 200, 88}
\definecolor{lightred}{RGB}{238, 102, 102}
\definecolor{lightblue}{RGB}{115, 192, 222}
\newtcolorbox{promptbox}[2][Prompt]{
colback=black!5!white,
arc=5pt, 
boxrule=0.5pt,
fonttitle=\bfseries,
title=#1, 
before upper={\small}, fontupper=\fontfamily{ptm}\selectfont,
colframe=#2,
}
\usepackage[utf8]{inputenc}
\usepackage{newunicodechar}

\newunicodechar{─}{\rule[0.5ex]{1em}{0.2ex}}
\newunicodechar{│}{\textbar}
\usepackage{verbatim}
\usepackage{microtype}
\usepackage{hyperref}
\usepackage{url}
\usepackage{booktabs}
\usepackage{enumitem}
\usepackage{graphicx}
\usepackage{multirow}
\usepackage{pifont}
\usepackage{xcolor}
\usepackage{booktabs}
\usepackage{hyperref}
\usepackage{makecell}
\usepackage{amsmath}
\usepackage{amsfonts}
\usepackage{enumitem}
\usepackage{multicol}
\usepackage{soul}
\usepackage{arydshln}
\usepackage{subcaption}
\usepackage{xspace}

\usepackage{caption}
\usepackage{color}  
\newcommand{\short}{\texttt{MAmmoTH2}\xspace}
\newcommand{\shortplus}{\texttt{MAmmoTH2-Plus}\xspace}

\newcommand{\model}[1]{\texttt{MAmmoTH2-#1}\xspace}
\newcommand{\modelplus}[1]{\texttt{MAmmoTH2-#1-Plus}\xspace}

\newcommand{\dataset}{\textsc{WebInstruct}\xspace}
\newcommand{\datasetplus}{\textsc{WebInstruct-Plus}\xspace}
\newcommand{\e}[1]{\underline{#1}}
\newcommand{\x}[1]{\textbf{#1}}

\title{MAmmoTH2: Scaling Instructions from the Web}

\author{%
  $^{\diamondsuit}$Xiang Yue\thanks{All of the authors are core contributors to the project.}~,
  $^{\spadesuit}$Tuney Zheng\footnotemark[1]~,
  $^{\spadesuit}$Ge Zhang\footnotemark[1]~,
  $^{\spadesuit}$Wenhu Chen\footnotemark[1] \\[5pt]
  $^\diamondsuit$Carnegie Mellon University, $^\spadesuit$University of Waterloo \\[3pt]
  \texttt{xyue2@andrew.cmu.edu ~~ wenhuchen@uwaterloo.ca } \\
}

\begin{document}

\maketitle
\vspace{-0.7cm}
\begin{center}
    \url{https://tiger-ai-lab.github.io/MAmmoTH2/}
\end{center}
\vspace{0.3cm}
\vspace{5pt}
\begin{abstract}
Instruction tuning improves the reasoning abilities of large language models (LLMs), with data quality and scalability being the crucial factors. Most instruction tuning data come from human crowd-sourcing or GPT-4 distillation. We propose a paradigm to efficiently harvest 10 million 
naturally existing instruction data from the pre-training web corpus to enhance LLM reasoning. Our approach involves (1) recalling relevant documents, (2) extracting instruction-response pairs, and (3) refining the extracted pairs using open-source LLMs.
Fine-tuning base LLMs on this dataset, we build \short models, which significantly boost performance on reasoning benchmarks. Notably, \model{7B}'s (Mistral) performance increases from 11\% to 36.7\% on MATH and from 36\% to 68.4\% on GSM8K without training on any in-domain data. Further training \short on public instruction tuning datasets yields \shortplus, achieving state-of-the-art performance on several reasoning and chatbot benchmarks.
Our work demonstrates how to harvest large-scale, high-quality instruction data without costly human annotation or GPT-4 distillation, providing a new paradigm for building better instruction tuning data.
\end{abstract}

\begin{figure}[!h]
    \centering
    \includegraphics[width=0.99\linewidth]{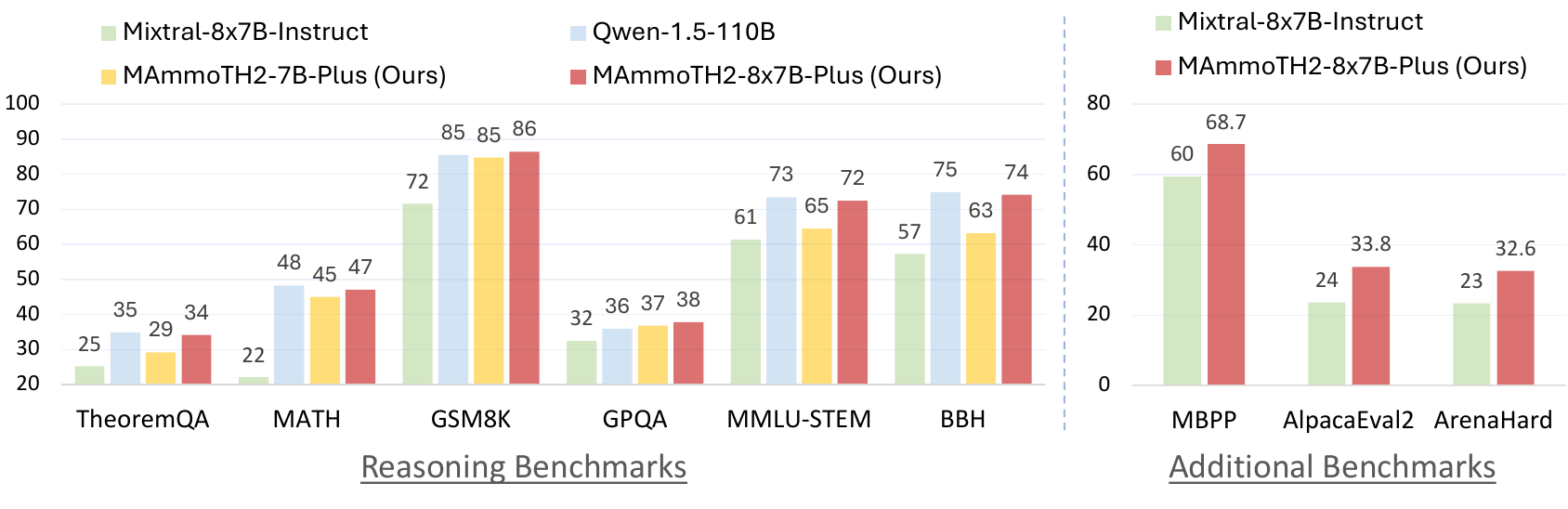}
    
    \caption{Overview of \texttt{MAmmoTH2-Plus} results. The \modelplus{8x7B} variant outperforms Mixtral-Instruct on reasoning benchmarks, matching Qwen-1.5-110B with only 13B active parameters. It also surpasses Mixtral-Instruct by around 10 points on general code and chatbot benchmarks.}
    \label{fig:front}
\end{figure}

\section{Introduction}
\input{introduction}

\section{\dataset}
\input{dataset}

\subsection{Additional Public Instruction Datasets}
\input{datasetplus}

\section{Experimental Setup}
\input{experiments}

\section{Conclusion}
In this paper, we argue that the web corpus contains a vast amount of high-quality instruction data across various domains. To mine this data, we develop a three-step pipeline consisting of recall, extraction, and refinement steps. Through this pipeline, we harvest \dataset, a total of 10M diverse, high-quality instruction-response pairs and train language models. Our experiments demonstrate that \short exhibits significantly enhanced science reasoning abilities compared to the baseline models. Our work showcases the potential of harnessing the vast amount of instruction data in the web corpus to democratize the development of LLMs with enhanced reasoning capabilities.


\input{bbl}
\clearpage
\appendix


\input{appendix}

\end{document}

%% file: introduction.tex
Reasoning is a fundamental aspect of human cognition and problem-solving~\citep{clark2018think,hendrycks2020measuring,cobbe2021training,rein2023gpqa,yue2023mmmu}. Proficiency in reasoning is essential for advancing scientific knowledge, developing new technologies, and making informed decisions in various contexts. 
Recently, large language models (LLMs)~\citep{brown2020language,ouyang2022training,touvron2023llama,touvron2023llama2,achiam2023gpt,team2023gemini} have shown remarkable progress in various NLP tasks. However, their ability to perform complex reasoning tasks~\citep{lin2024criticbench} in the domains of mathematics, science, and engineering is still limited.

Recent studies have extensively explored how to enhance base LLMs' reasoning abilities. 
The two main approaches are continued training and instruction tuning. Continued training trains LLMs on large-scale filtered documents~\citep{lewkowycz2022solving,taylor2022galactica,azerbayev2023llemma,shao2024deepseekmath,ying2024internlm}. 
Instruction tuning seeks to employ supervised fine-tuning loss on, usually small-scale, high-quality instruction-response pairs~\citep{ouyang2022training,chung2024scaling}. While human-annotated instruction datasets~\citep{cobbe2021training,hendrycks2021measuring,amini2019mathqa} are often limited in scale, recent studies~\citep{yu2023metamath,yue2023mammoth,toshniwal2024openmathinstruct,li2024common,tang2024mathscale} attempt to prompt GPT-4 with seed data to increase the scalability. However, the synthesized instruction data becomes highly biased, not diverse, and prone to a high degree of hallucination.

\begin{figure}[!t]
    \centering
    \includegraphics[width=0.92\linewidth]{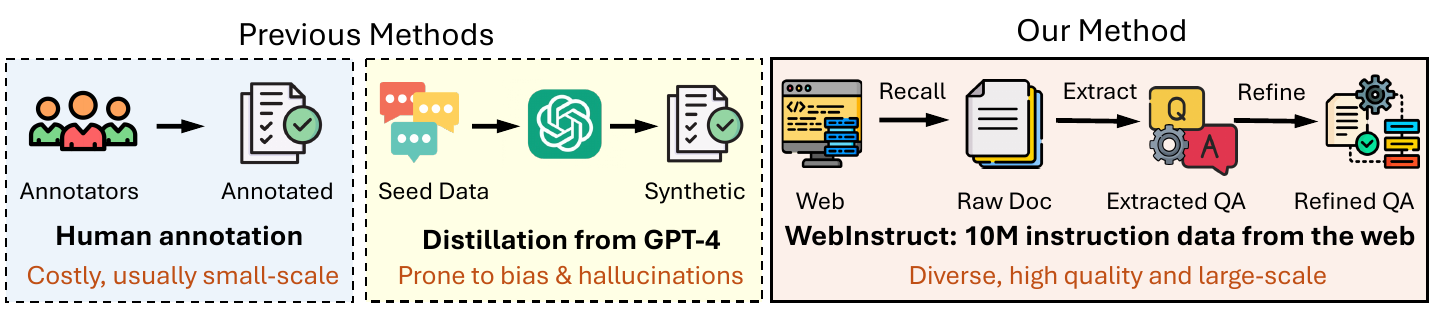}    
    \caption{Comparison between our dataset curation method and previous studies.}
    \label{fig:scaling_up}
    \vspace{-4ex}
\end{figure}

To address these limitations, we propose to \textbf{discover naturally existing instruction data from the web} (\autoref{fig:scaling_up}). We argue that the pre-training corpus (e.g., Common Crawl) already contains a vast amount of high-quality instruction data for LLM reasoning. 
For example, the web corpus contains a large amount of educational materials in the form of instruction-following pairs. These documents range across various domains like math, science, engineering, and humanities. Such readily available instruction data is not only diverse but also of high quality. However, such instruction data is highly dispersed across the corpus, which makes it particularly challenging to discover.

In this paper, we aim to mine these instruction-response pairs from the web using a three-step pipeline. \textbf{(1) Recall step}: We create a diverse seed dataset by crawling several quiz websites. We use this seed data to train a fastText model~\citep{joulin2016fasttext} and employ it to recall documents from Common Crawl~\citep{together2023redpajama}. GPT-4 is used to trim down the recalled documents by their root URL. We obtain 18M documents through this step. \textbf{(2) Extract step}: We utilize open-source LLMs like Mixtral~\citep{jiang2024mixtral} to extract Q-A pairs from these documents, producing roughly 5M candidate Q-A pairs. \textbf{(3) Refine step}: After extraction, we further employ Mixtral-8$\times$7B~\citep{jiang2024mixtral} and Qwen-72B~\citep{bai2023qwen} to refine~\citep{zheng2024kun} these candidate Q-A pairs. This refinement operation aims to remove unrelated content, fix formality, and add missing explanations to the candidate Q-A pairs. This refinement operation is pivotal to maintaining the quality of the mined Q-A pairs. Eventually, we harvest a total of 10M instruction-response pairs through these steps. Unlike existing instruction-tuning dataset, our dataset \dataset is purely mined from the Web without any human crowdsourcing or GPT-4 distillation.

We validate the effectiveness of \dataset by training \short on various base models (\autoref{fig:front}), including Mistral-7B~\citep{jiang2023mistral}, Llama3-8B~\citep{MetaLLama3_2024}, Mixtral-8$\times$7B~\citep{jiang2024mixtral}, and Yi-34B~\citep{young2024yi}. \short significantly outperforms the base models on seven held-out reasoning benchmarks: TheoremQA~\citep{chen2023theoremqa}, GSM8K~\citep{cobbe2021training}, MATH~\citep{hendrycks2021measuring}, ARC-C~\citep{clark2018think}, MMLU-STEM~\citep{hendrycks2021measuring}, GPQA~\citep{rein2023gpqa}, and BBH~\citep{suzgun2022challenging}. \model{7B} improves Mistral-7B's performance by an average of 14 absolute points, while \model{34B} enhances Yi-34B's performance by an average of 5.8 absolute points. Notably, Mistral-7B's MATH accuracy can rise from 11.2\% to 36.7\% after training on \dataset. As our dataset contains no in-domain data from our evaluation benchmarks, this highlights the models' strong generalization ability.

We further enhance \short's performance on code generation, math reasoning, and instruction-following tasks by tuning it on open-source instruction datasets, including OpenHermes2.5~\citep{OpenHermes}, Code-Feedback~\citep{zheng2024opencodeinterpreter}, and Math-plus. The resulting model, \shortplus, excels on seven reasoning benchmarks and other general tasks. \modelplus{7B} and \modelplus{8B} achieve state-of-the-art performance on TheoremQA, ARC-C, MMLU-STEM, GPQA, and BBH, and competitive results on MATH (45\%) and GSM8K (85\%). \shortplus also performs well on general tasks, with \modelplus{7B} showing promising results on HumanEval and MBPP, and \model{8×7B} leading the AlpacaEval 2.0 and Arena Hard leaderboards. 

Interestingly, \modelplus{8B} and Llama-3-8B-Instruct, both tuned from Llama-3-base using datasets of the same size (10M), provide an apple-to-apple comparison. The only distinction is that Llama-3-8B-Instruct is trained on 10M human-annotated dataset while we do not require any human annotation. \modelplus{8B} outperforms Llama-3-Instruct by 6 points on reasoning tasks while matching its performance on general tasks, reflecting \dataset's cost-effectiveness advantage. \shortplus consistently surpasses official instruction models like Mixtral-Instruct on chat benchmarks. These results demonstrate the effectiveness of our approach to scale up instruction data from the web and offer a new perspective for future instruction tuning studies.

%% file: dataset.tex
\begin{figure}[!b]
    \centering
    \includegraphics[width=1.0\linewidth]{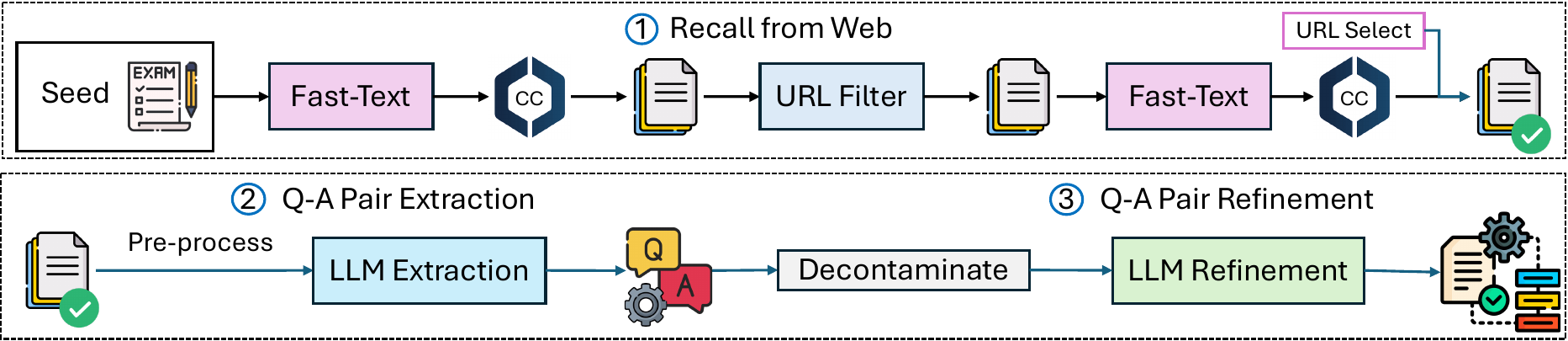}
    \caption{Step 1: Recall relevant documents from Common Crawl. Step 2: Extracting Q-A pairs. Step 3: Refine with the extracted Q-A pairs.}
    \label{fig:recall}
\end{figure}
\begin{figure}[!b]
    \centering
    \includegraphics[width=1.0\linewidth]{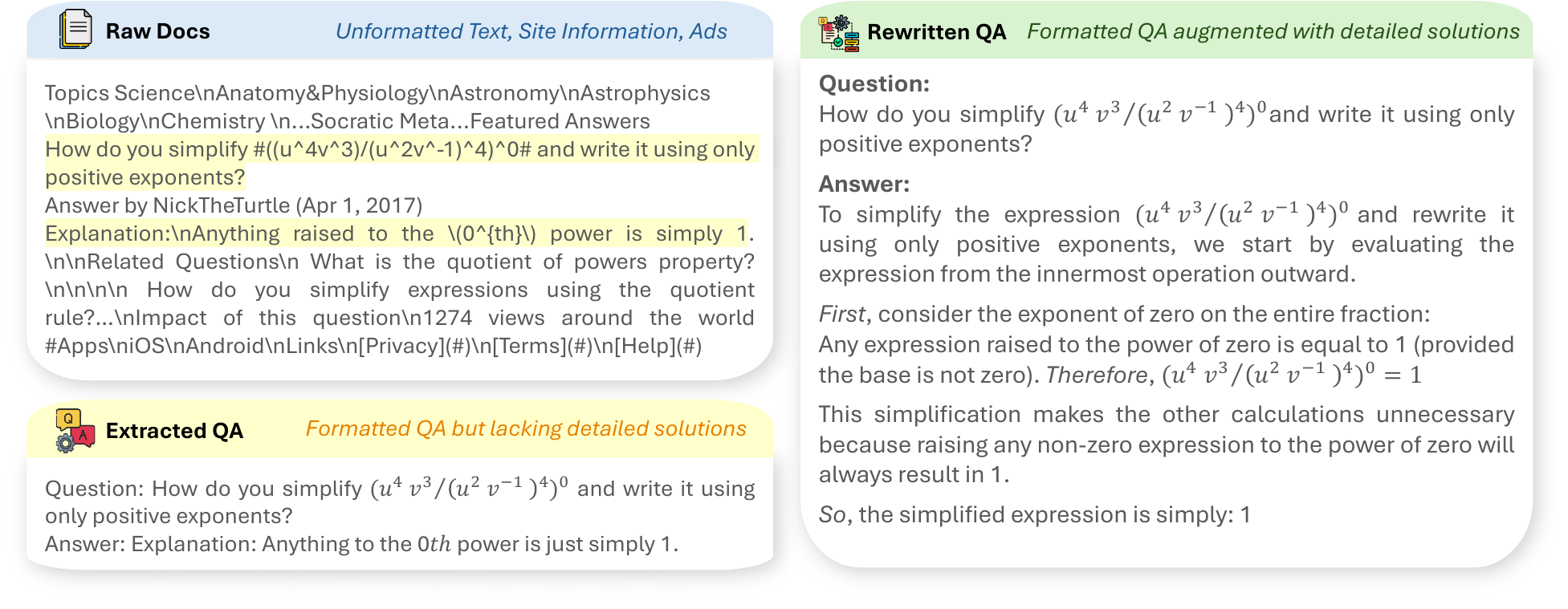}
    \caption{An illustrating example from \dataset for the extraction and refinement step.}
    \label{fig:refine}
\end{figure}

In this section, we outline the process of constructing \dataset. Specifically, we divide the data collection pipeline into three stages: (1) relevant document recall from the web corpus, (2) Q-A pair extraction from recalled document, and (3) Q-A pair refinement. The full pipeline is depicted in~\autoref{fig:recall} and an example for extraction and refinement is provided in~\autoref{fig:refine}.

\subsection{Recall from Common Crawl}
In contrast to previous math-centric approaches~\citep{paster2023openwebmath,wang2023generative,shao2024deepseekmath}, we aim for broad coverage of disciplines such as math, science, engineering, etc. Therefore, careful balancing of the seed data is necessary to ensure diversity. However, publicly available training datasets are mostly limited to mathematics. To address this issue, we propose to crawl new exam problems from several educational websites. These sites contain diverse problems from various disciplines, helping to ensure diversity.
We crawled 100K seed data as positive training examples and randomly selected 100K negative documents from CC~\citep{together2023redpajama} to train a fastText model~\citep{joulin2016fasttext}. 
The trained fastText model is used to recall relevant documents. 
We employ the open-source fastText library with a vector dimension of 256 to train the model for 3 epochs, with a learning rate of 0.1, a maximum n-gram length of 3, and a maximum number of word occurrences of 3. 
We recalled 100B tokens using the trained fasttext model from an internal CC.
These raw web documents are further grouped by their domains (root URL) and only domains with more than 1000 documents are retained.
We extracted roughly 600K domains from the recalled documents.
We then prompt GPT-3.5 to scan through the domains and automatically select those that might contain instruction data.
Around 50K domains are further labeled as positive samples by GPT-3.5.
Note that all the recalled documents in the first round are \textit{not kept} for further usage in Q-A Pair Extraction and Refinement.
Next, we sample documents from the selected domains as positive examples, and documents from the non-selected domains and general Common Crawl as negative examples to re-train an improved fastText classifier. The newly trained fastText classifier is used to recall documents.
We recalled 40B tokens using the newly trained fastText model.
We prompt GPT-4 to sift through the recalled domains again, ultimately leading to 18M raw documents, primarily originating from the desired websites. 

\subsection{Q-A Pair Extraction}
We observe that a significant number of naturally existing Q-A pairs are present in the 18M documents. However, these Q-A pairs are interspersed with a high volume of noise such as ads, markups, boilerplate, etc. Our preliminary training on these raw documents only yields limited gains.

First, we carefully pre-process the HTML to pre-extract useful content from the recalled documents. This is mostly rule-based filtering to clean site information, ads, HTML boilerplate, etc. This step significantly reduces the document length for the next stage. We then prompt Qwen-72B~\citep{bai2023qwen} to identify the question and answer pairs from the preprocessed documents. Specifically, we provide a few in-context examples to help the model understand what to extract. We also allow the model to return void if no natural question-answer pairs exist. In this stage, only 30\% of the recalled documents were identified as containing naturally existing Q-A pairs, resulting in roughly 5M Q-A pairs as our candidates for the next step. However, these candidates still contain a substantial amount of unrelated content and formality issues. Besides that, a large portion of the extracted Q-A pairs also lack explanations for how the answer is derived. Therefore, we propose to perform another round of refinement to increase the data quality.

To avoid contamination, we follow previous work~\citep{shao2024deepseekmath} and filter out web pages containing questions or answers to all of our evaluation benchmarks. Specifically, we filter out all web pages that contain $n$-grams ($n=10$) string matches with either the questions or answers.

\subsection{Q-A Pair Refinement}
To further improve the extracted Q-A pair candidates, we propose refining them using LLMs. In this step, we prompt Mixtral-22B$\times$8~\citep{jiang2024mixtral} and Qwen-72B~\citep{bai2023qwen} to reformat the extracted Q-A pairs. If the answer does not contain any explanation, these two LLMs will attempt to complete the intermediate reasoning steps leading to the given answer. We adopt two models to increase the diversity of our dataset. Eventually, we harvest 10M Q-A pairs as our final instruction-tuning dataset \dataset.

\subsection{Dataset Statistics}
To better distinguish our dataset from the existing ones, we include a summarization table in~\autoref{tab:datasets}. It can be observed that most SFT datasets contain less than 1M samples but are of high quality. XwinMath~\citep{li2024common} is the largest dataset, scaling up to over 1M samples through GPT4 synthesis, while OpenMathInstruct~\citep{toshniwal2024openmathinstruct} has not been generated using GPT-4 but instead uses Mixtral-8x7B~\cite{jiang2024mixtral}. However, the seed data for both datasets is only based on GSM and MATH, leading to narrow domain coverage. In contrast, continue-training (CT) datasets are normally filtered from the web with much larger size, often exceeding 10B tokens and even rising to 120B tokens. However, continued pre-training on these massive datasets can be not only expensive but also ineffective due to the high noise ratio. \dataset, with roughly 5B tokens, strikes a good balance between scalability and quality. It approaches the scalability of common CT datasets while maintaining high quality through the three-step construction pipeline. This makes our dataset unique compared to other alternatives. 
\input{Tables_and_Figures/datasets}

%% file: Tables_and_Figures/datasets.tex
\begin{table}[!t]
\small
\centering
\caption{The list of existing supervise-fine-tuning (SFT) and continue-training (CT) datasets. SFT datasets are primarily from academic NLP sources or synthesized by GPT-3.5/4 using seed data. CT datasets are larger but nosier. Our dataset falls between these two types.}
\resizebox{\linewidth}{!}{
\begin{tabular}{lllll}
\toprule
Dataset       & \#Pairs & Domain  & Format & Dataset Source                         \\
\midrule
FLAN V2~\citep{chung2024scaling}       & 100K & General & SFT   & NLP data + Human CoT          \\
Self-Instruct~\citep{wang2023self} & 82K  & General & SFT   & Generated by GPT3                         \\
GPT4-Alpaca~\citep{alpaca}   & 52K  & General & SFT   & Generated by GPT4                         \\
SuperNI~\citep{wang2022super}       & 96K  & General & SFT   & NLP Datasets  \\
Tora~\citep{gou2023tora}          & 16K  & Math    & SFT   & GSM+MATH Synthesis by GPT4           \\
WizardMath~\citep{luo2023wizardmath}    & 96K  & Math    & SFT   & GSM+MATH Synthesis by GPT4         \\
MathInstruct~\citep{yue2023mammoth}  & 262K & Math    & SFT   & Math datasets Synthesis by GPT4        \\
MetaMathQA~\citep{yu2023metamath}    & 395K & Math    & SFT   & GSM+MATH Synthesis by GPT3.5         \\
XwinMath~\citep{li2024common}      & 1.4M & Math    & SFT   & GSM+MATH Synthesis by GPT4       \\
OpenMathInstruct~\citep{toshniwal2024openmathinstruct} & 1.8M  & Math  & SFT & GSM+MATH Synthesis by Mixtral      \\
\midrule
Dataset       & \#Tokens & Domain  & Format & Dataset Source                        \\
\midrule
OpenWebMath~\citep{paster2023openwebmath}   & 12B    & Math  & LM    & Filtered from Web       \\
MathPile~\citep{wang2023generative}      & 10B    & Math  & LM    & Filtered from Web       \\
Cosmopeida~\citep{benallal2024cosmopedia}    & 25B    & General & LM  & Synthesized by Mixtral   \\
MINERVA~\citep{lewkowycz2022solving}       & 38B    & Math  & LM    & Filtered from Web        \\
Proof-Pile-2~\citep{azerbayev2023llemma}  & 55B    & Math  & LM    & OpenWebMath+Arxiv+Code \\
Galactica~\citep{taylor2022galactica}     & 106B   & Math \& Sci. & LM & Filtered from Web  \\
DeepseekMath~\citep{shao2024deepseekmath}  & 120B   & Math  & LM    & Recalled from Web        \\
\midrule
\dataset       & (10M) 5B       & Math \& Sci.  & SFT  &  Recall and Extracted from Web \\   
\bottomrule
\end{tabular}
}
\label{tab:datasets}
\vspace{-3ex}
\end{table}

%% file: datasetplus.tex
To further enhance the diversity and quality of our dataset, we fine-tune \short on several open-source instruction tuning datasets. These datasets are carefully selected based on their relevance to different reasoning subjects. Additionally, we consider some chat datasets to balance reasoning ability and general chat ability. The open-source datasets we incorporate are {OpenHermes 2.5}~\citep{OpenHermes}, {Code-Feedback}~\citep{zheng2024opencodeinterpreter} and our Math-Plus, which is an augmented version of MetaMathQA (395K)~\citep{yu2023metamath} and Orca-Math (200K)~\citep{mitra2024orcamath}. More details of the public datasets can be found in Appendix \ref{sec:additional_datasets}.

%% file: experiments.tex
\input{Tables_and_Figures/main_table}

\subsection{Training Setup}
We unify all the samples in our instruction dataset to conform to the structure of a multi-turn instruction tuning dataset. This standardization ensures that the fine-tuned models can process data consistently, regardless of the original dataset formats. We select the open-source models Mistral 7B~\citep{jiang2023mistral}, Mixtral 8$\times$7B~\citep{jiang2024mixtral}, Llama-3 8B~\citep{MetaLLama3_2024}, and Yi-34B~\citep{young2024yi} as our base models. We fine-tune these models to validate our \dataset at multiple scales using the LLaMA-Factory~\citep{zheng2024llamafactory} library. We use a learning rate of 5e-6 for Mistral 7B and 1e-5 for Mixtral, Llama-3 8B, and Yi 34B. The global batch size is set to 512 with a maximum sequence length of 4096. We employ a cosine scheduler with a 3\% warm-up period for 2 epochs. To efficiently train the models, we utilize DeepSpeed~\citep{rasley2020deepspeed} with the ZeRO-3 stage. All the models are trained with 32 A100 GPUs. 

\subsection{Evaluation Datasets}

\label{sec:eval_dataset}

To rigorously assess the capabilities of models in reasoning abilities across different domains, we utilize several widely used datasets, GSM8K~\citep{cobbe2021training}, MATH~\citep{hendrycks2021measuring}, TheoremQA~\citep{chen2023theoremqa}, {BIG-Bench Hard (BBH)}~\citep{suzgun2022challenging}, {ARC-C}~\citep{clark2018think}, {GPQA}~\citep{rein2023gpqa}, {MMLU-STEM}~\citep{hendrycks2020measuring}. These datasets collectively enable a comprehensive assessment of language models' reasoning prowess across a spectrum of complexity and realism. The details of the evaluation datasets are in Appendix \ref{sec:eval_dataset_details}.

We further evaluate the models on additional code generation tasks (including HumanEval~\citep{chen2021evaluating}, MBPP~\citep{austin2021program} and their augmented version~\citep{liu2024your}), general LLM benchmarks like MMLU~\citep{hendrycks2020measuring} and its recent robust and challenging version MMLU-Pro~\citep{MMLU-Pro}. We also consider chat benchmarks like MT-Bench~\citep{zheng2024judging}, AlpacaEval 2.0~\citep{alpaca_eval}, and Arena Hard~\citep{arenahard2024} to demonstrate the generalizability of \dataset and \datasetplus on more general LLM benchmarks.

\section{Main Results}
\subsection{Experimental Results on Reasoning Benchmarks}
\autoref{table:results} presents our main results, with existing models partitioned into two tracks based on their parameter size. For 7B parameter models, we observe that our model trained solely with \dataset achieves significant improvements over the base models. For instance, \model{7B} boosts the performance of Mistral-7B by an average of 14 points. Notably, \dataset does not contain any training data from these evaluation benchmarks, making all evaluations essentially held-out. The substantial performance gains demonstrate the strong generalization capabilities of \model{7B}. Similarly, \model{8B} boosts the performance of Llama-3-8B-base by an average of 8.8 points. We also experiment with larger models like Yi-34B and Mixtral to show that the performance gains are consistent across the board. Notably, Yi-34B's performance on MATH also increases by 19\% after training on \dataset.

Further tuning on several additional public datasets also significantly enhances performance. The \shortplus model family achieves state-of-the-art results across the board. For example, \shortplus's performance on TheoremQA, GPQA, and ARC-C represents the best-known results for any model under 10B parameters. \modelplus{7B}'s performance on MATH and GSM is also close to the best-known results. We also show the results of the models solely trained on the additional public datasets in  \autoref{sec:impact_additional_public_data}.  

An interesting comparison is between \modelplus{8B} and Llama3-Instruct, as both models are trained from the Llama3-base. Llama-3-instruct was trained on a 10M human-annotated instruction dataset along with public datasets, similar to \dataset combined with additional public datasets. Therefore, these two models are highly comparable. Our experiments show that \modelplus{8B} outperforms Llama3-Instruct by an average of 6\% across the benchmarks. This substantial gain indicates that \dataset is highly cost-effective. For larger models, we found that \modelplus{8x7B} can even match the performance of Qwen-1.5-110B with only 13B active parameters. These results demonstrate the effectiveness of our scalable instruction tuning approach.

\subsection{Additional Experimental Results}
To further demonstrate the capabilities of our models beyond the reasoning benchmarks presented in~\autoref{table:results}, we conduct additional experiments to evaluate their performance on code generation, general language understanding, and instruction-following tasks. \autoref{tab:chat_results} showcases the results of various models on code generation tasks. The \modelplus{7B} model exhibits strong performance, achieving the highest average scores of 66.1 and 58.2 on HumanEval(+) and MBPP(+) datasets, respectively. It outperforms the official instruct counterparts like Mistral-7B-Instruct-v0.2 on these metrics, indicating its superior code generation abilities.

\begin{figure}[!b]
\centering
\vspace{-20pt}
\includegraphics[width=\linewidth]{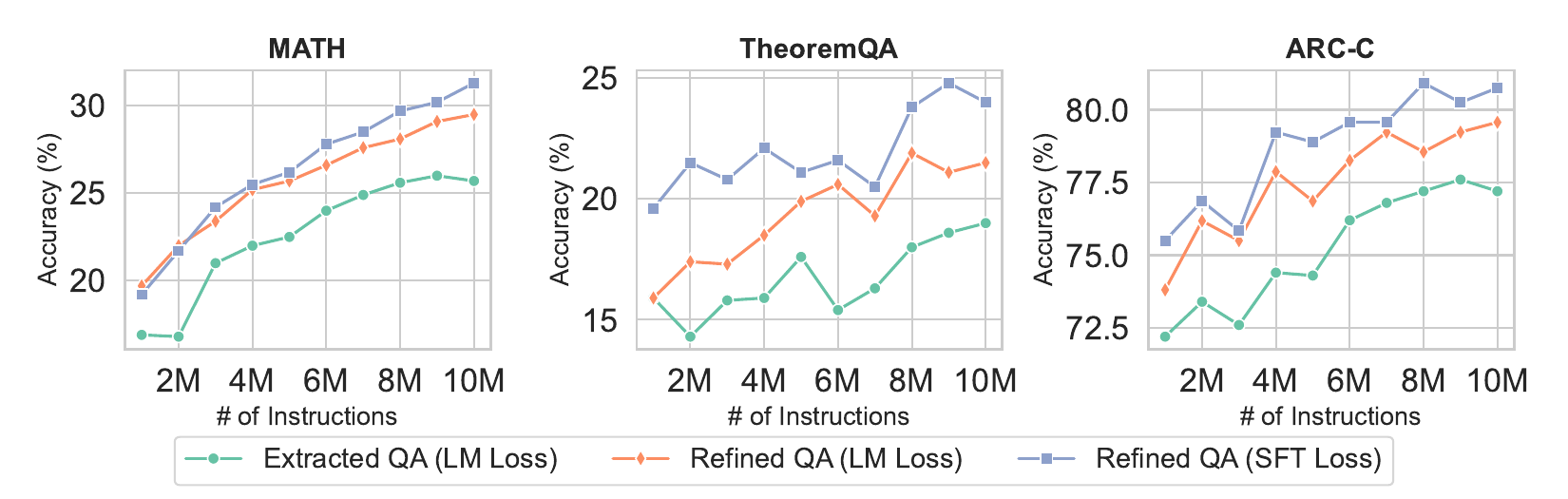}
\caption{Mistral-7B model reasoning performance improves with scaling instructions. Additionally, SFT Loss is a more effective learning approach compared to LM Loss.}
\label{fig:scaling_effect}
\end{figure}

To assess the general language understanding and instruction-following capabilities of our models, we evaluate them on a range of benchmarks, as shown in \autoref{tab:chat_results}. The \shortplus models exhibit strong performance across these tasks, showcasing their versatility and robustness. For example, \modelplus{8×7B} achieves the highest scores on AlpacaEval 2.0 and Arena Hard leaderboards, surpassing competitive models like GPT-3.5-Turbo and Tulu-2-DPO-70B~\citep{ivison2023camels}.

The strong performance of \short on code generation and general language understanding tasks, as evidenced by \autoref{tab:chat_results}, demonstrates that our method does not overfit to the reasoning benchmarks. Instead, it shows the models' ability to generalize well to a wide range of tasks, highlighting their versatility and robustness. These additional experiments further validate the effectiveness of our \dataset in developing powerful and flexible language models.

\input{Tables_and_Figures/additional_table}

\section{Ablation Study}

\subsection{Scaling Effect of Instructions}
We first investigate the impact of model scaling and loss functions on the performance of language models across three representative tasks: MATH, TheoremQA, and ARC-C. We train models with varying training samples (1M to 10M) using extracted QA and refined QA data, and compare the effectiveness of two training losses: LM loss and SFT loss. \autoref{fig:scaling_effect} shows that increasing model size and using SFT Loss with synthetic data consistently improves accuracy across all tasks. These findings demonstrate the importance of model scaling and supervised fine-tuning with synthetic data for enhancing language model performance in various domains.

\begin{wraptable}{r}{8cm} 
\centering
\vspace{-15pt}
\caption{Comparison of the two data-refining LLMs. We train the three models with the same steps. }
\small
\begin{tabular}{@{}lccccc@{}}
\toprule
Data & GSM & MATH & MMLU-S & Theo. & ARC. \\ \midrule
Mixtral & 62.9 & 29.1 & 56.5 & \textbf{26.1} & 78.3 \\
 Qwen & 65.4 & 28.9 & \textbf{60.6} & 23.5 & 80.8 \\\midrule
\textbf{Merged} & \textbf{65.6} & \textbf{31.0} & 60.5 & 24.8 & \textbf{81.8} \\ \bottomrule
\end{tabular}%
\label{tab:my-table}
\end{wraptable}

\subsection{Comparison of Two Refined Models} To assess the effectiveness of the Q-A pair refinement process by different LLMs, we conducted experiments by training three mistral-7B models: one on the data refined by Mixtral-22B$\times$8, another on the data refined by Qwen-72B, and a third on the merged samples refined by both models. For a fair comparison, we trained the models with the same 9000 steps and a global batch size of 512. Our results show that the model trained on Mixtral-22B$\times$8 refined data achieves comparable performance to the one trained on Qwen-72B refined data. The model trained on the merged samples consistently outperforms the models trained on data refined by individual LLMs. This demonstrates the effectiveness of using multiple LLMs for refinement, as it leads to a more diverse and comprehensive dataset.

\subsection{Comparison of Different Domains and Sources.} To understand how each domain (e.g., math, science, others) and data source (e.g., forum websites and education websites) contribute to the training, we train Mistral 7B on the subsets of different domains and data sources. Details of how we obtain domain labels can be found in ~\autoref{sec:category}.

\begin{wraptable}{r}{8cm} 
\vspace{-10pt}
\centering
\caption{Impact of different data domains and sources.}
\small
\begin{tabular}{@{}lccccc@{}}
\toprule
Data Source & GSM & MATH & MMLU-S & Theo. & ARC. \\ \midrule
Base & 47.4 & 15.7 & 51.4 & 17.3 & 77.6 \\ \midrule
Forum & 51.0 & 24.0 & \textbf{54.7} & 21.0 & 78.2 \\
Education & \textbf{58.0} & \textbf{24.8} & 54.3 & \textbf{23.2} & \textbf{79.5} \\ \midrule
Math & 52.9 & \textbf{27.3} & 51.6 & \textbf{21.7} & 74.1 \\
Science & 54.4 & 23.7 & \textbf{58.9} & 21.0 & \textbf{83.6} \\
Other & \textbf{59.4} & 20.8 & 55.3 & 21.1 & 79.4 \\ \bottomrule
\end{tabular}%
\label{tab:data_source}
\end{wraptable}

As shown in \autoref{tab:data_source}, training on different domains and data sources leads to varied performance across the evaluation benchmarks. The education website data source consistently outperforms the forum data source, indicating the higher quality of educational questions. Interestingly, while the math domain excels on MATH, it does not lead to significant improvements on GSM8K, another math-focused dataset, suggesting that training on a single math dataset may not generalize well to other math benchmarks. Furthermore, training solely on the math domain does not yield substantial gains on science and STEM benchmarks, highlighting the need for diverse training. In contrast, the "Other" domain, which includes a diverse range of subjects, achieves the highest score on GSM8K, emphasizing the importance of diversity in the training data.

\subsection{Case Study}
\begin{wrapfigure}{r}{0.4\textwidth}
\vspace{-17pt}
  \centering
\includegraphics[width=0.4\textwidth]{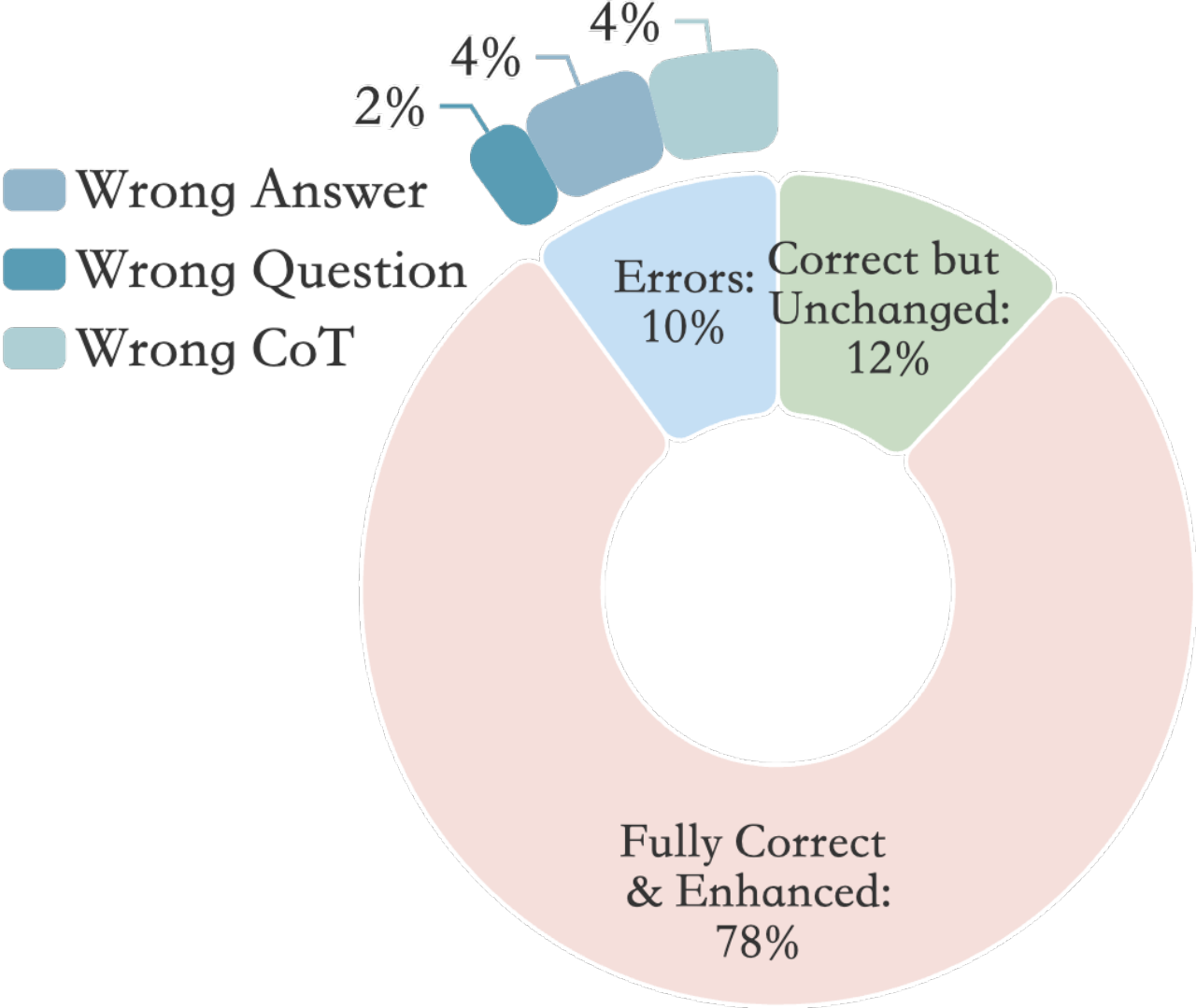}
  \caption{Quality distribution of 50 sampled refined QA examples.}
  \vspace{-5pt}
  \label{fig:error_analysis}
\end{wrapfigure}
We further conduct a case study examining the quality of extracted and refined QA pairs from the dataset. We showcase some good and bad cases in~\autoref{sec:case_study}. We observe that the question/answer pairs extracted from well-formed exam and education websites are of high quality. The common issue is that a large portion of extracted answers do not contain intermediate rationale (chain-of-thought). This issue could lead to worse generalization. 

Therefore, we prompt Mixtral and Qwen-72B to complete the intermediate steps. We observe that the success rate of such completion is relatively high. However, there are cases where the extracted question/answer pairs contain serious formatting issues, which pose challenges for the following refinement step. Besides these issues, we also observe that LLMs can sometimes modify the intention of the originally extracted content, causing hallucinations.

To quantify the error percentages, we randomly sample 50 refined QA examples and ask the human annotators to compare whether the refined examples are correct and significantly better than the extracted ones in terms of format and intermediate solutions. As we can see from ~\autoref{fig:error_analysis}, 78\% examples have been improved after refinement and only 10\% examples introduce hallucinations after refinement.  Overall, our case study reveals that the harvested instruction tuning dataset is generally accurate with a low error rate.

%% file: Tables_and_Figures/main_table.tex
\begin{table}[!t]
\caption{Main results on reasoning datasets. Models without the '-Instruct' suffix refer to the released base models. Results are taken from official papers or blogs when available; otherwise, we use our own evaluation script. Underscored results represent the best baseline scores under the size constraint. All models are inferred with few-shot CoT: TheoremQA (5-shot), MATH (4-shot), GSM8K (4-shot), GPQA (5-shot), MMLU-STEM (5-shot), BBH (3-shot), and ARC-C (8-shot).}
\small
\centering
\resizebox{\linewidth}{!}{
\begin{tabular}{lcccccccc}
\toprule
Model                 & TheoremQA & MATH                         & GSM8K                        & GPQA & MMLU-ST &BBH & ARC-C & AVG \\
\midrule
GPT-4-Turbo-0409      & 48.4      & 69.2                         & 94.5                         & 46.2 & 76.5   & 86.7 & 93.6 & 73.6   \\
\midrule
\multicolumn{9}{c}{\textbf{Parameter Size between 20B and 110B}} \\
\midrule
Qwen-1.5-110B         & \e{34.9}  & \e{49.6}                     & \e{85.4}                     & 35.9 & \e{73.4}  & \e{74.8} & 91.6 & \e{63.6} \\
Qwen-1.5-72B          & 29.3      & 46.8                         & 77.6                         & \e{36.3} & 68.5  & 68.0 & \e{92.2} &  59.8 \\
Deepseek-LM-67B       & 25.3      & 15.9                         & 66.5                         & 31.8 & 57.4   & 71.7 & 86.8 & 50.7 \\
\e{Yi-34B}            & 23.2      & 15.9                         & 67.9                         & 29.7 & 62.6   & 66.4 & 89.5 & 50.7 \\
Llemma-34B            & 21.1      & 25.0                         & 71.9                         & 29.2 & 54.7   & 48.4 & 69.5 & 45.7  \\
\e{Mixtral-8$\times$7B}& 23.2     & 28.4                         & 74.4                         & 29.7 & 59.7   & 66.8 & 84.7 & 52.4  \\
Mixtral-8$\times$7B-Instruct &  25.3     &  22.1                 & 71.7                         & 32.4 & 61.4   & 57.3 & 84.7 & 50.7  \\
Intern-Math-20B       & 17.1      & 37.7                         & 82.9                         & 28.9 & 50.1   & 39.3 & 68.6 & 46.4  \\
\midrule
\multicolumn{9}{c}{Trained only with \dataset (All evaluations are held-out)} \\
\midrule
\model{34B}           & 30.4    & 35.0                         & 75.6                         & 31.8 & 64.5   & 68.0  & 90.0  & 56.4 \\
\rowcolor{LightCyan}
$\Delta$ over \e{Yi}  & +7.2    & +19.1                        & +7.7                         & +2.1 & +2.9   & +1.2  & +0.5  & +5.8 \\
\model{8x7B}          & 32.2    & 39.0                         & 75.4                         & 36.8 & 67.4   & 71.1  & 87.5  & 58.9 \\
\rowcolor{LightCyan}
$\Delta$ over \e{Mixtral}& +9.2   & +10.6                      & +1.0                         & +7.1 & +7.4   & +3.3  & +2.8  & +6.5 \\
\midrule
\multicolumn{9}{c}{Continue trained with additional instruction datasets (All held-out except MATH and GSM8K)} \\
\midrule
\modelplus{8x7B}  &  \x{34.1}    & \x{47.0}                    & \x{86.4}                     & \x{37.8} & \x{72.4} & \x{74.1}  & \x{88.4}  & \x{62.9} \\
\rowcolor{LightCyan}
$\Delta$ over Qwen-1.5-110B & -0.8      & -2.6                 & +1.0                         & +1.5     & -1.0   & -0.7  & -4.0  & -0.7 \\
\midrule
\multicolumn{9}{c}{\textbf{Parameter Size = 7B or 8B}} \\
\midrule
Deepseek-7B           & 15.7      & 6.4                        & 17.4                         & 25.7 &  43.1  & 42.8 & 47.8  & 28.4  \\
Qwen-1.5-7B           & 14.2      & 13.3                       & 54.1                         & 26.7 &  45.4  & 45.2 & 75.6  & 39.2 \\
\e{Mistral-7B}        & 19.2      & 11.2                       & 36.2                         & 24.7 &  50.1  & 55.7 & 74.2  & 38.8 \\
Gemma-7B              & 21.5      & 24.3                       & 46.4                         & 25.7 &  53.3  & 57.4 & 72.5  & 43.0  \\
Llemma-7B             & 17.2      & 18.0                       & 36.4                         & 23.2 &  45.2  & 44.9 & 50.5  & 33.6 \\
WizardMath-7B-1.1     & 11.7      & 33.0                       & \e{83.2}                     & 28.7 &  52.7  & 56.7 & 76.9  & 49.0 \\
Abel-7B-002           & 19.3      & 29.5                       & 83.2                         & 30.3 &  29.7  & 32.7 & 72.5  & 42.5 \\
Intern-Math-7B        & 13.2      & 34.6                       & 78.1                         & 22.7 &  41.1  & 48.1 & 59.8  & 42.5  \\
Rho-1-Math-7B         & 21.0      & 31.0                       & 66.9                         & 29.2 &  53.1  & 57.7 & 72.7  & 47.3  \\
Deepseek-Math-7B      & \e{25.3}  & 34.0                       & 64.2                         & 29.2 &  56.4  & 59.5 & 67.8  & 48.0 \\
Deepseek-Math-Instruct& 23.7      & \e{44.3}                   & 82.9                         & 31.8 &  59.3  & 55.4 & 70.1  & 52.5 \\
\e{Llama-3-8B}        & 20.1      & 21.3                       & 54.8                         & 27.2 &  55.6  & 61.1 & 78.6  & 45.5 \\
Llama-3-8B-Instruct   & 22.8      & 30.0                       & 79.5                         & \e{34.5} &  \e{60.2}  & \e{66.0} & \e{80.8} & \e{53.4}  \\
\midrule
\multicolumn{9}{c}{Trained only with \dataset (All evaluations are held-out)} \\
\midrule
\model{7B}            & 29.0      & 36.7                       & 68.4                         & 32.4  & 62.4  & 58.6  & 81.7  & 52.8 \\
\rowcolor{LightCyan}
$\Delta$ over \e{Mistral} & +9.8  & +25.5                      & +32.2                        & +7.7  & +12.3 & +2.9  & +7.5  & +14.0 \\
\model{8B}            & 32.2      & 35.8                       & 70.4                         & 35.2  & 64.2  & 62.1  & 82.2  & 54.3  \\
\rowcolor{LightCyan}
$\Delta$ over \e{Llama3}  & +12.2 & +14.5                      & +15.6                        & +8.0  & +8.6  & +1.0  & +3.6  & +8.8  \\
\midrule
\multicolumn{9}{c}{Continue trained with additional instruction datasets (All held-out except MATH and GSM8K)} \\
\midrule
\modelplus{7B}        &  29.2     & \x{45.0}                  &  \x{84.7}                    &  36.8  & 64.5     &  63.1     & 83.0      & 58.0     \\
\modelplus{8B}        &  \x{32.5} & 42.8                      &  84.1                        &\x{37.3}& \x{65.7} & \x{67.8}  & \x{83.4}  & \x{59.1} \\
\rowcolor{LightCyan}
$\Delta$ over best baseline & +7.2   & +0.7                   & +1.5                         & +2.8   & +5.5     & +1.8      & +2.6  & +5.7 \\
\bottomrule
\end{tabular}
}
\vspace{-20pt}
\label{table:results}
\end{table}

%% file: Tables_and_Figures/additional_table.tex
\begin{table}[!t]
\caption{Evaluation of code generation, instruction-following and MMLU(-Pro) performance for various models.  We report the average of HumanEval(+) and MBPP (+) accuracy as the code generation performance (breakdown results are in ~\autoref{sec:code_gen}). Baseline scores are sourced from the original papers or the EvalPlus, MT-Bench, AlpacaEval 2.0, Arena Hard and MMLU-Pro leaderboards. (``-'') indicates that the score is not available from the sources. \shortplus exhibits strong general conversational ability and excels at multitask language understanding across a wide range of domains compared to their official instruct counterparts and larger models.}
\small
\centering
\begin{tabular}{@{}lcccccc@{}}
\toprule
 & \begin{tabular}[c]{@{}c@{}}Code \\ Generation\end{tabular} & MT-Bench & \begin{tabular}[c]{@{}c@{}}Alpaca\\ Eval 2.0\end{tabular} & \begin{tabular}[c]{@{}c@{}}Arena\\ Hard\end{tabular} & MMLU & MMLU-Pro \\ \midrule
GPT-4-1106-preview & 85.6 (77.5) & 9.32 & 50.0 & - & - & - \\
GPT-3.5-Turbo-1106 & 79.7 (70.2) & 8.32 & 19.3 & 18.9 & - & - \\
GPT-3.5-Turbo-0301 & - & 7.94 & 18.1 & 18.1 & 70.0 & - \\ \midrule
Tulu-2-DPO-70B & 51.2 (43.0) & 7.89 & 21.2 & 15.0 & 67.8 & 40.5 \\
Llama-2-70b-chat & 31.4 (26.5) & 6.86 & 14.7 & 11.6 & 63.0 & 33.6 \\
Yi-34B-Chat & 38.7 (32.6) & 7.86 & 27.2 & 23.1 & 73.5 & 42.1 \\ \midrule
Mistral-7B-Instruct-v0.2 & 43.4 (36.5) & 7.60 & 17.1 & 12.6 & 60.8 & 30.8 \\
Llama-3-8B-Instruct & \e{65.8} (\e{58.0}) & 8.02 & 22.9 & 20.6 & 67.2 & 40.9 \\
Mixtral-8$\times$7B-Instruct-v0.1 & 52.3 (44.7) & \textbf{8.30} & \e{23.7} & \e{23.4} & \textbf{70.6} & 41.0 \\ \midrule
\modelplus{7B} & \textbf{66.1} (\textbf{58.2}) & 7.88 & 23.4 & 14.6 & 63.3 & 40.9 \\
\modelplus{8B} & 61.9 (53.3) & 7.95 & 18.5 & 16.6 & 64.6 & \e{43.4} \\
\modelplus{8x7B} & 63.3 (55.3) & \e{8.20} & \textbf{33.8} & \textbf{32.6} & \e{68.3} & \x{50.4} \\ \bottomrule
\end{tabular}%
\label{tab:chat_results}
\vspace{-10pt}
\end{table}

%% file: appendix.tex
\section{Details of Additional Public Instruction Tuning Datasets}
\label{sec:additional_datasets}

\begin{itemize}[leftmargin=*]

\item \textbf{OpenHermes 2.5}~\citep{OpenHermes}: The OpenHermes-2.5 dataset is a comprehensive collection of diverse data sources for instruction tuning, including 1M examples from datasets in math, science, and coding, alongside synthetic and chat-based data. It incorporates diverse sources such as Airoboros 2.2, CamelAI Domain Expert Datasets, ChatBot Arena, Collective Cognition, Evol Instruct, Glaive Code Assistant, GPT4-LLM, GPTeacher, Medical Tasks, MetaMath, SlimOrca, Platypus, ShareGPT, and Unnatural Instructions GPT4. We remove TheoremQA from Platypus as it is one of our test sets.

\item \textbf{Code-Feedback}~\citep{zheng2024opencodeinterpreter}: The Code-Feedback dataset is a multi-turn code generation and refinement dataset, containing 68,000 multi-turn interactions between users, code generation models, and compiler systems. It includes initial user instructions followed by compiler and user feedback. This dataset significantly enhances the model's multi-turn interaction coding ability.

\item \textbf{Math-Plus}: This dataset combines public datasets such as MetaMathQA (395K)~\citep{yu2023metamath} and Orca-Math (200K)~\citep{mitra2024orcamath}. Both of these datasets are generated by GPT-3.5/4 using GSM/MATH and other math datasets as the seed data. To further augment the dataset, we prompt GPT-4 to rewrite Q-A pairs from MATH training sets, adding an additional 300K examples to enhance the challenging problems. The total size of the Math-Plus dataset is 894K examples.

\end{itemize}

To ensure consistency and compatibility, we carefully align the format and structure of these additional datasets with \dataset. These supplementary datasets provide a rich resource for training models to answer questions and provide explanations across a wide range of topics, enhancing their versatility and applicability in real-world scenarios.

\section{Details of Evaluation Datasets}

\label{sec:eval_dataset_details}

To rigorously assess the capabilities of models in reasoning abilities across different domains, we utilize several widely used datasets. Each of these datasets is designed to challenge the models in different aspects of reasoning.

\begin{itemize}[leftmargin=*]

\item \textbf{GSM8K}~\citep{cobbe2021training}: This test dataset contains 1.32K diverse grade school math problems, intended to test basic arithmetic and reasoning ability in an educational context.

\item \textbf{MATH}~\citep{hendrycks2021measuring}: Comprising 5000 intricate competition-level problems to evaluate the models' ability to perform complex mathematical reasoning.

\item \textbf{TheoremQA}~\citep{chen2023theoremqa}: Focused on applying mathematical theorems to solve advanced problems in fields such as mathematics, physics, and engineering, TheoremQA includes 800 questions that test the theoretical reasoning capabilities.

\item \textbf{BIG-Bench Hard (BBH)}~\citep{suzgun2022challenging}: Consisting of 23 tasks previously found challenging for language models from BIG-Bench~\citep{srivastava2023beyond}, BBH contains a total of 6511 challenging problems examining the capability of LLMs to solve them.

\item \textbf{ARC-C}~\citep{clark2018think}: ARC includes questions derived from various grade-level science exams, testing models' ability to handle both straightforward and complex scientific queries. We use the challenge subset, which contains 1,172 test questions.

\item \textbf{GPQA}~\citep{rein2023gpqa}: This dataset provides "Google-proof" questions in biology, physics, and chemistry, designed to test deep domain expertise and reasoning under challenging conditions. We use the diamond subset containing 198 hard problems.

\item \textbf{MMLU-STEM}~\citep{hendrycks2020measuring}: Spanning 57 subjects across multiple disciplines, MMLU evaluates the breadth and depth of a model's knowledge in a manner akin to academic and professional testing environments. We select the STEM subset of MMLU with 3.13K problems.

\end{itemize}

These datasets collectively enable a comprehensive assessment of language models' reasoning prowess across a spectrum of complexity and realism. We further evaluate the models on additional code generation tasks (including HumanEval~\citep{chen2021evaluating}, MBPP~\citep{austin2021program} and their augmented version~\citep{liu2024your}), general LLM benchmarks like MMLU~\citep{hendrycks2020measuring}, and chat benchmarks like MT-Bench~\citep{zheng2024judging}, AlpacaEval 2.0~\citep{alpaca_eval}, and Arena Hard~\citep{arenahard2024} to demonstrate the generalizability of \dataset and \datasetplus on more general LLM benchmarks.

\section{Related Work}
\input{related_work}

\newpage
\section{Code Generation Results}
\label{sec:code_gen}
We report the code generation results of our models and baselines in \autoref{tab:code_results}.

\begin{table}[!h]
\small
\centering
\begin{tabular}{@{}lcccccc@{}}
\toprule
& HumanEval &HumanEval+ & MBPP &MBPP+ &Average &Average+ \\ \midrule
Mistral-7B & 28.7 &23.8 &51.9 &42.1 &40.3 &33.0 \\
Gemma-7B & 26.8 &20.1 & 52.6 &43.4 &39.7 &31.8 \\
Llama-3-8B &33.5 &29.3 & 61.4 &51.6 &47.5 &40.5 \\ \midrule
Gemma-1.1-7B-Instruct & 42.7 &35.4 & 57.1 &45.0 &49.9 &40.2 \\
Mistral-7B-Instruct-v0.2 & \textbf{75.0} &\textbf{70.1} &44.7 &37.0 &59.9 &53.6 \\
Llama-3-8B-Instruct & 61.6 &56.7 & \textbf{70.1} &\textbf{59.3} &\e{65.9} &\e{58.0} \\ 
Mixtral-8$\times$7B-Instruct-v0.1 & 45.1 &39.6 &59.5 &49.7 &52.3 &44.7 \\\midrule
\model-7B-Plus & \e{72.1} & \e{65.9} & 60.1 &50.4 &\textbf{66.1} &\textbf{58.2} \\
\model-8B-Plus & 63.4 &57.9 & 60.4 &48.6 &61.9 &53.3 \\
\model-8$\times$7B-Plus & 57.9 &53.7 & \e{68.7} &\e{56.9} &63.3 &55.3 \\\bottomrule
\end{tabular}%
\caption{Code generation results of different models. Baseline results are copied from the EvalPlus~\citep{liu2024your} leaderboard.}
\label{tab:code_results}
\end{table}

\section{Impact of Additional Public Instruction Tuning Datasets}
\label{sec:impact_additional_public_data}
In \autoref{sec:additional_datasets}, we introduce additional public instruction tuning datasets to further boost the model's reasoning performance. Here, we show the three setups of models trained on: 1) \dataset only; 2) Additional Public Datasets only; 3) \dataset + Additional Public Datasets (which we first trained on \dataset and then continued training on additional public datasets). The results are shown in \autoref{tab:ablation_webinstruct}.

\begin{table}[!h]
\small
\centering

\resizebox{\linewidth}{!}{
\begin{tabular}{lcccccccc}
\toprule
Data                 & TheoremQA & MATH                         & GSM8K                        & GPQA & MMLU-ST &BBH & ARC-C & AVG \\
\midrule
\multicolumn{9}{c}{Mistral 7B Base} \\ \midrule
\dataset             & 29.0      & 36.7                       & 68.4                         & 32.4  & 62.4  & 58.6  & 81.7  & 52.8 \\
\textsc{Public Datasets} & 22.6 &37.9 & 83.5 & 29.3 & 57.6 &62.7 &79.9&53.4 \\
\textsc{WebIns.+Public.}        &  \x{29.2}     & \x{45.0}                  &  \x{84.7}                    &  \x{36.8}  & \x{64.5}     &  \x{63.1}     & \x{83.0}      & \x{58.0}     \\ \midrule
\multicolumn{9}{c}{Mixtral 8x7B Base} \\ \midrule

\dataset          & 32.2    & 39.0                         & 75.4                         & 36.8 & 67.4   & 71.1  & 87.5  & 58.9  \\
\textsc{Public Datasets} &31.3  &45.1 &85.3 &37.4 &69.4 &73.2 &88.1 &61.4\\ 
\textsc{WebIns.+Public.}  &  \x{34.1}    & \x{47.0}                    & \x{86.4}                     & \x{37.8} & \x{72.4} & \x{74.1}  & \x{88.4}  & \x{62.9} \\
\bottomrule
\end{tabular}
}
\caption{Impact of additional public instruction tuning datasets. }
\label{tab:ablation_webinstruct}
\end{table}

\section{Distributions of Website Domains in \dataset}
\begin{figure}[!h]
    \centering
    \includegraphics[width=0.8\linewidth]{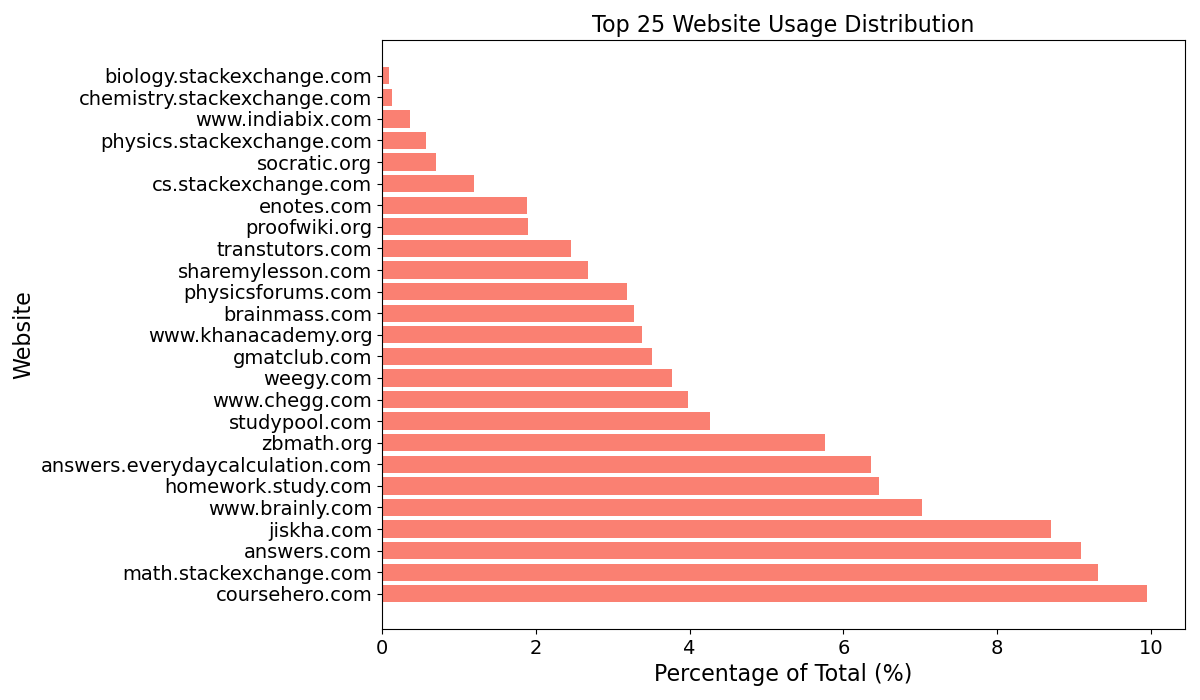}
    \caption{The distribution of the top websites in our instruction dataset.}
    \label{fig:urls}
\end{figure}

\autoref{fig:urls} show the distribution of the top websites in \dataset. 

\newpage
\section{Domain Distribution of \dataset}
\label{sec:category}
\begin{figure}[!h]
    \centering
    \includegraphics[width=0.75\linewidth]{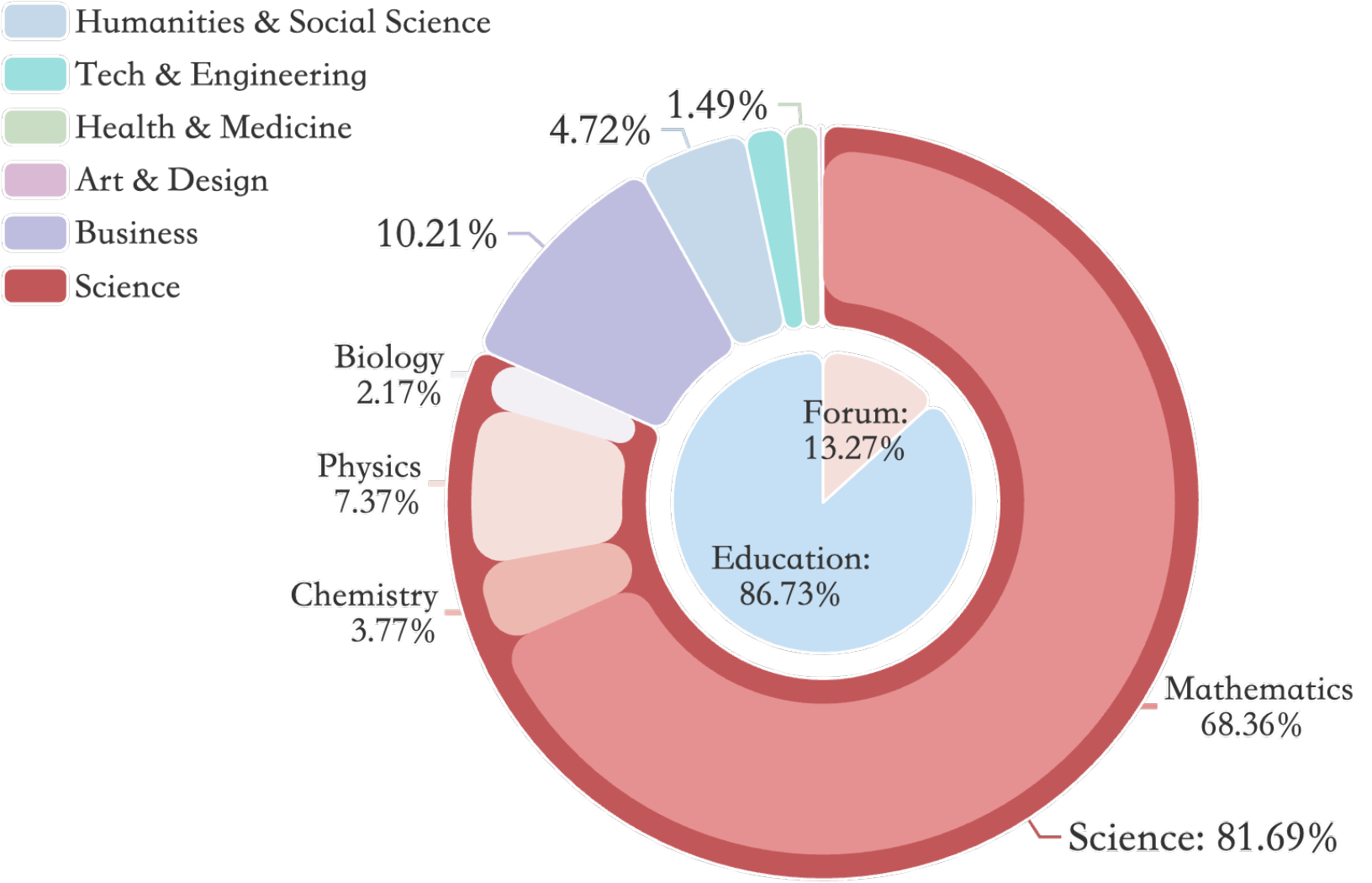}
    
    \caption{ Breakdown of \dataset by subject domains and data sources. }
    \label{fig:subject_distribution}
\end{figure}

\autoref{fig:subject_distribution} presents a breakdown of the \dataset by subject domains and data sources, providing insights into the composition and diversity of the mined instruction-response pairs. The subject labels are automatically annotated using the Llama-3-8B-Instruct model, while the distribution between education and forum data is obtained by analyzing the source URLs of the samples. The pie chart reveals that \dataset is predominantly composed of science-related subjects, with 81.69\% of the data falling under the broad "Science" category. Within this category, Mathematics takes up the largest share at 68.36\%, followed by Physics, Chemistry, and Biology. This highlights the dataset's strong emphasis on mathematical problem-solving and scientific reasoning. The remaining non-science categories, such as Business, Art \& Design, and Health \& Medicine, contribute to the diversity of the dataset. In terms of data sources, the vast majority (86.73\%) of the instruction-response pairs come from exam-style questions, while forum discussions make up the remaining 13.27\%. This source breakdown indicates that \dataset primarily consists of well-structured, educational content, supplemented by real-world discussions and inquiries from forums. The diverse subject coverage and the combination of education and forum data enable \dataset to capture a wide range of reasoning tasks and problem-solving scenarios.

\section{Limitations of \dataset}
\label{sec:limitation}
Despite employing a three-step pipeline to ensure the quality of the mined instruction-response pairs, there may still be some noise and inaccuracies in the dataset, as mentioned in \autoref{fig:error_analysis}. The extraction and refinement steps rely on the performance of the LLMs used, which may introduce biases and errors. Future work could explore more advanced techniques for data cleaning and validation, such as human-in-the-loop approaches or training a data selection model for filtering. Furthermore, while \dataset covers a wide range of subjects, including math, science, and engineering, there may be specific subdomains or advanced topics that are underrepresented, such as humanities and other daily chat topics. Expanding the coverage of the seed data to include broader and more diverse instruction-response pairs could further enhance the reasoning capabilities of LLMs in different areas.

\section{Broader Social Impact}
\label{sec:broader_impact}
The development of \texttt{MAmmoTH2} and the \dataset has the potential to positively impact education by providing the community with access to a large-scale, diverse set of instruction-response pairs across various subjects, particularly in mathematics and science. \texttt{MAmmoTH2} can assist students in their learning process by offering step-by-step solutions, explanations, and targeted feedback, thus enhancing their understanding of complex concepts and problem-solving skills. Furthermore, the dataset and models can serve as valuable resources for educators to create more effective and engaging educational content, assignments, and assessments.
However, it is important to consider potential limitations and biases in the dataset and the model. The quality and accuracy of the instruction-response pairs in \dataset may vary, and the model's outputs may reflect these inconsistencies. Therefore, it is crucial to employ rigorous data filtering and quality control measures to ensure the reliability and trustworthiness of the dataset and the model. 

\newpage
\section{Case Study}

\renewcommand{\thetable}{A\arabic{table}} 
\setcounter{table}{0}
\renewcommand{\thefigure}{A\arabic{figure}} %
\setcounter{figure}{0}

\label{sec:case_study}
In this section, we aim to review some of the instances in \dataset. The data extracted from the website contains many issues. For example, the most common issue is that the answer is relatively short without any intermediate rationale, which could occur frequently from the quiz website with multi-choice questions. Another common issue is that the extraction destroys important formatting in the document, which leads to unrecoverable errors.   

\begin{promptbox}[Good Case 1]{lightgreen}
\textbf{Extracted:} \\
\texttt{Question:} Find the distance from the point ${\frac{3}{2}, \frac{3}{2}}$ to the line $-\frac{37 x}{10}-\frac{21 y}{10}-\frac{7}{2}=0$. \\
\texttt{Answer:} 61 $\sqrt{\frac{2}{905}}$. \\

\textbf{Refined:} \\
\texttt{Question:} Calculate the distance from the point ($\frac{3}{2}, \frac{3}{2}$) to the line defined by the equation $-\frac{37}{10}x - \frac{21}{10}y - \frac{7}{2} = 0$. \\

\texttt{Answer:} To find the distance 'd' from a point $(x_0, y_0)$ to a line given by $Ax + By + C = 0$, we use the formula:
$$
d = \frac{|Ax_0 + By_0 + C|}{\sqrt{A^2 + B^2}}.
$$
For the point $\left(\frac{3}{2}, \frac{3}{2}\right)$ and the line $-\frac{37}{10}x - \frac{21}{10}y - \frac{7}{2} = 0$, substitute the coordinates into the formula:
$$
d = \frac{|-\frac{37}{10} \cdot \frac{3}{2} - \frac{21}{10} \cdot \frac{3}{2} - \frac{7}{2}|}{\sqrt{\left(-\frac{37}{10}\right)^2 + \left(-\frac{21}{10}\right)^2}}.
$$
Simplifying further, the exact simplified distance 'd' is  $61 \sqrt{\frac{2}{905}}$. Thus, the correct distance from the point to the line is $61 \sqrt{\frac{2}{905}}$. 
\end{promptbox}
\vspace{-0.5cm}
\captionof{figure}{In this Refined version, the expression of both the question and the answer is clearer and more precise. It employs standardized mathematical symbols and terms, such as representing the coordinates of the point and the equation of the line as $(x_0, y_0)$ and $Ax + By + C = 0$, respectively. Moreover, it explicitly provides the formula for calculating the distance. Through these revisions, readers can more easily understand how to compute the distance from a point to a line, making the solution process more transparent and easier to follow.}

\vspace{0.75cm}

\begin{promptbox}[Good Case 2]{lightgreen}
\textbf{Extracted:} \\
\texttt{Question:} What are the two main ways that minerals are classified? \\
\texttt{Answer:} They can be classified by luster, streak, breakage, and hardness. \\

\textbf{Refined:} \\
\texttt{Question:} What are the primary criteria used for classifying minerals in mineralogy? \\
\texttt{Answer:} Minerals are primarily classified based on two main categories: physical properties and chemical properties. The physical properties include luster, streak, crystal form, hardness, density, and cleavage or breakage. Chemical properties involve the mineral's chemical composition and reactions with other substances. These characteristics help identify and distinguish different minerals in mineralogy.
\end{promptbox}
\vspace{-0.5cm}
\captionof{figure}{Refining the extracted response provides a more comprehensive and accurate explanation of mineral classification in mineralogy. By expanding on the initial response to include both physical and chemical properties, the refined version offers a deeper understanding of how minerals are categorized and identified. This enriched content can enhance the model's ability to generate informative responses related to mineralogy.}

\begin{promptbox}[Good Case 3]{lightgreen}
\textbf{Extracted:} \\
\texttt{Question:} How do you find theasymptotes for $Q(x)=(2x^2)/(x^2 -5x -6)$? \\
\texttt{Answer:} vertical  asymptotes x=-1, x=6/n horizontal asymptote y=2 /n /n /n /n /n /n /n /n /n /n \\

\textbf{Refined:} \\
\texttt{Question:} How do you determine the vertical and horizontal asymptotes for the function \( Q(x) = \frac{2x^2}{x^2 - 5x - 6} \)? \\
\texttt{Answer:} To find the vertical asymptotes, we need to identify the values of \( x \) that make the denominator equal to zero:

\[
x^2 - 5x - 6 = (x - 6)(x + 1) = 0
\]

This gives us the vertical asymptotes at \( x = -1 \) and \( x = 6 \).

For the horizontal asymptote, as \( x \) approaches infinity or negative infinity, the dominant terms in the numerator and denominator are both \( x^2 \). Thus, we compare their coefficients:

- Numerator's highest degree coefficient: \( 2 \)
- Denominator's highest degree coefficient: \( 1 \)

The horizontal asymptote is given by the ratio of these coefficients:

\[
y = \frac{2}{1} = 2
\]

Therefore, the vertical asymptotes are at \( x = -1 \) and \( x = 6 \), and the horizontal asymptote is \( y = 2 \).

\end{promptbox}
\vspace{-0.4cm}
\captionof{figure}{The refined answer provides a more detailed explanation for finding the asymptotes. It includes the steps for solving for the vertical asymptotes by factoring the denominator and determining where it equals zero. It also explains how to find the horizontal asymptote by comparing the leading coefficients of the highest-degree terms in both the numerator and the denominator. The use of LaTeX formatting makes the mathematical notation clearer and easier to understand.}

\vspace{0.9cm}

\begin{promptbox}[Good Case 4]{lightgreen}
\textbf{Extracted:} \\
\texttt{Question:}Which of the following is not an example of biased language? a. “You are too short to reach that.” b. “Are you deaf?” c. “Move along, grandpa!” d. “I think I’ve been gypped. \\
\texttt{Answer:} D. I think I've been gypped. \\

\textbf{Refined:} \\
\texttt{Question:} Which of the following statements does not exhibit biased language?

\begin{itemize}
    \item[a.] ``You are too short to reach that.''
    \item[b.] ``Are you deaf?''
    \item[c.] ``Move along, grandpa!''
    \item[d.] ``I think I've been gypped.''
\end{itemize}

\texttt{Answer:} The statement that does not exhibit biased language is:

\begin{itemize}
    \item [d.] "I think I've been gypped."
\end{itemize}

Explanation: Biased language refers to words or phrases that stereotype, discriminate, or show prejudice towards a particular group. In this case, options a, b, and c contain biases towards people's physical abilities (height), hearing ability, and age, respectively. However, "I think I've been gypped" is an informal expression meaning to be cheated or deceived, which, while potentially offensive, does not directly target a specific group based on attributes like race, gender, or disabilities.
\end{promptbox}
\vspace{-0.4cm}
\captionof{figure}{The refined version of the input presents the question and answers in a clearer, more structured format that improves readability and comprehension. Specifically, using a bulleted list for the options makes them distinct and easier to evaluate individually. Moreover, the addition of an explanation provides context and justification for the selected answer, enhancing the educational value of the content. This refined structure and the inclusion of an explanatory note make it a superior training example for models, as it helps in understanding not just the answer but the rationale behind it, promoting deeper learning and better pattern recognition in language processing tasks.}

\begin{promptbox}[Good Case 5]{lightgreen}
\textbf{Extracted:} \\
\texttt{Question:} I am stucked on the following challenge: "If the line determined by two distinct points $(x_1, y_1)$ and $(x_2, y_2)$ is not vertical, and therefore has slope $(y_2-y_1)/(x_2-x_1)$, show that the point-slope form of its equation is the same regardless of which point is used as the given point." Okay, we can separate $(x_0, y_0)$ from the form to get: $$y(x_2-x_1)-x(y_2-y_1) = y_0(x_2-x_1)-x_0(y_2-y_1)$$ But how exclude this point $(x_0,y_0)$ and leave only $x, y, x_1, y_1, x_2, y_2$ in the equation? UPDATE: There is a solution for this challenge: $$(y_1-y_2)x+(x_2-x_1)y=x_2y_1-x_1y_2$$ From the answer I found that $$y_2(x-x_1)-y_1(x-x_2)=y(x_2-x_1)$$ ... but why this is true? \\
\texttt{Answer:} Thanks to saulspatz, the solution is to simply show that, whether we are using $(x_1, y_1)$ or $(x_2, y_2)$ as the given point, the equation does not change. 

So both equations:
\[
y - y_1 = m (x - x_1)
\]
\[
y - y_2 = m (x - x_2)
\]
reduce to:
\[
(y_1 - y_2)x + (x_2 - x_1)y = x_2y_1 - x_1y_2.
\]

\textbf{Refined:} \\
\texttt{Question:} Demonstrate that the point-slope form of the equation of a line determined by two distinct points \((x_1, y_1)\) and \((x_2, y_2)\), where the line is not vertical, remains consistent regardless of which point is chosen as the reference. Why does the form of the equation not change when different reference points are used? \\
\texttt{Answer:} To address this challenge, we need to confirm that the point-slope equation of a line remains consistent regardless of the point chosen to derive it. Given two points \((x_1, y_1)\) and \((x_2, y_2)\) with a non-vertical line passing through them, the slope \(m\) of the line is \(\frac{y_2-y_1}{x_2-x_1}\).

The point-slope form of a line's equation based on point \((x_1, y_1)\) is:
$$y - y_1 = m(x - x_1)$$

Similarly, using point \((x_2, y_2)\) gives:
$$y - y_2 = m(x - x_2)$$

These can be rewritten and compared to show their equivalence. Expanding both equations:
- From the first equation: \(y - y_1 = \frac{y_2-y_1}{x_2-x_1}(x - x_1)\)
- From the second equation: \(y - y_2 = \frac{y_2-y_1}{x_2-x_1}(x - x_2)\)

We simplify these to find a common form. Multiplying both sides by \((x_2 - x_1)\) and rearranging terms, we find:
$$y_2(x - x_1) - y_1(x - x_2) = y(x_2 - x_1)$$

This confirms that both forms collapse to the same linear equation, showing the consistency of the point-slope form irrespective of the chosen reference point. Thus, it is demonstrated that the structural form of the equation remains unchanged, ensuring that the mathematical properties of the line—defined by its slope and points through which it passes—are invariant to the choice of the reference point used in its expression. This uniformity stems from the linearity and the constant slope defined by any two points on a straight line.

\end{promptbox}
\captionof{figure}{The original answer contains an irrelevant introductory sentence, which makes it less focused. In contrast, the refined version provides a clearer question by explicitly stating the problem of demonstrating the invariance of the point-slope form. The explanation in the answer is expanded to show step-by-step equivalence between the two forms and includes detailed reasoning about why the consistency holds. The added explanation reinforces the understanding of linear equations and the concept of constant slope, making the solution more comprehensive and pedagogically valuable.}

\newpage
\begin{promptbox}[Bad Case 1]{lightred}
\textbf{Extracted:} \\
\texttt{Question:} \text{Sort} $\left\{-3\pi, 9, \frac{17}{\sqrt{3}}, -4, \frac{47}{7}, \frac{3}{e}, \frac{3}{\sqrt{5}}\right\}$.\\
\texttt{Answer:} $\left\{-3\pi, -4, \frac{3}{e}, \frac{3}{\sqrt{5}}, \frac{47}{7}, 9, \frac{17}{\sqrt{3}}\right\}$.

\textbf{Refined:} \\
\texttt{Question:} Sort the following numbers in ascending order: $-3\pi, 9, \frac{17}{\sqrt{3}}, -4, \frac{47}{7}, \frac{3}{e}, \frac{3}{\sqrt{5}}$. \\
\texttt{Answer:} The sorted numbers in ascending order are: $-3\pi, -4, \frac{3}{e}, \frac{3}{\sqrt{5}}, \frac{47}{7}, 9, \frac{17}{\sqrt{3}}$.
\end{promptbox}
\vspace{-0.5cm}
\captionof{figure}{The refined example does not enhance the clarity or quality of the information. The original formatting and question structure were already clear and concise, and the transformation into a slightly more verbose question-and-answer format does not add significant value. Additionally, both versions use proper LaTeX formatting for mathematical expressions, so there is no improvement in readability or technical accuracy. Essentially, the refinement introduces additional words without enhancing comprehension, which could potentially dilute the effectiveness of the model training by focusing on unnecessary stylistic alterations rather than content clarity or quality.}

\vspace{1.2cm}

\begin{promptbox}[Bad Case 2]{lightred}
\textbf{Extracted:} \\
\texttt{Question:}  Which curve (AD or SRAS) will be affected and which direction (leftward or rightward) will it shift when Congress goes to war against Canada?\\
\texttt{Answer:} When Congress falls into a war with Canada, the aggregate demand curve will be affected by moving to the left. Suppose Congress fights against Canada, then, it is likely that Congress will pass laws that advocate for increased tariffs on imports.\\ Spending on exports is an element of aggregate demand. Thus, its curve shifts rightwards, when its components, which are, spending on consumptions, government spending, or spending on investment increase.\\ On the other hand, Canada will not do business with the USA as usual when they fall out with Congress. For example, the expenses on exports to Canada will reduce.\\ Thus, the aggregate demand will fail to increase and shift on the left if there is less consumption from households, falls of investments from firms or when the demand for exports from a country like Canada falls.\\

\textbf{Refined:} \\
\texttt{Question:} How does a war between Congress and Canada affect the Aggregate Demand (AD) curve, and in which direction will it shift? \
\texttt{Answer:} When Congress declares war against Canada, the Aggregate Demand (AD) curve shifts to the left because Congress will implement increased tariffs on imports. Spending on exports, which is a component of aggregate demand, decreases due to these tariffs.\ The Short Run Aggregate Supply (SRAS) curve, however, is not directly affected by this situation. It mainly reflects changes in production costs and the availability of resources, which are not directly influenced by international conflicts unless they lead to disruptions in the domestic production process.\ In summary, the AD curve shifts left due to decreased exports and potentially lower consumer and business spending, while the SRAS curve remains relatively stable unless there are indirect impacts on domestic production. \\

\end{promptbox}
\vspace{-0.4cm}
\captionof{figure}{The refined answer, while more structured and clearer in terms of the specific economic curves discussed, introduces inaccuracies that were not present in the original. Specifically, the refined version inaccurately states that spending on exports decreases due to tariffs, which contradicts the fundamental economic concept that tariffs affect imports rather than exports. Furthermore, the original answer provides a broader analysis of potential economic impacts, such as changes in household consumption and firm investment, which are omitted in the refined version. These omissions reduce the depth and accuracy of the economic analysis provided in response to the effects of war on economic curves.}

\begin{promptbox}[Bad Case 3]{lightred}
\textbf{Extracted:} \\
\texttt{Question:} Is there a tool (LaTeX, JavaScript, Mathematica..) that allows one to draw Venn diagram automatically from subsets relations, e.g. A $\subset A+B, A \subset C, C \subset C+D, B \not\subset C$ would yield:

\begin{verbatim}
────────────────
|               |
|       A+B     | 
|               |
────────────────
|               |
────────────────
────────────────
|               |
|               |
|       A       |
|               |
|               |
────────────────
────────────────
|               |
────────────────────────────
|             C+D          |
|                          |
────────────────────────────
\end{verbatim}
(Sorry for the pitiful Ascii drawing) \\

\texttt{Answer:}https://tex.stackexchange.com/questions/9681/how-to-draw-venn-diagrams-especially-complement-s-in-latex answers it in latex but I couldn't get their answer to work in my answer, as it used tikzpicture ( seems unsupported here).\\

\textbf{Refined:} \\
\texttt{Question:} Is there any software for generating diagrams to show relationships among set members, for example, \( A \subset A+B \)? \\
\texttt{Answer:} One might try using LaTeX, which supports drawing with the TikZ package, although some features may not work in certain settings. For more details, check out this link: https://tex.stackexchange.com/questions/9681/how-to-draw-venn-diagrams-especially-complement-s-in-latex. If problems persist, updating the software or consulting the community might help.

\end{promptbox}
\captionof{figure}{The refined example's quality becomes bad because it omits most of the specific subset relations provided in the original query, making it unclear which relationships need to be visualized. The answer shifts the focus towards potential troubleshooting rather than directly addressing the user's need to generate diagrams based on specific subset conditions. Moreover, the original question included a sample ASCII diagram for clarity, which is entirely dropped in the refined version, removing a helpful virtual context.}

%% file: related_work.tex
\paragraph{Instruction Tuning.}
Instruction tuning is crucial for aligning large language models (LLMs) with end tasks. There are two main types of instruction tuning data: (1) human-written data, such as FLAN~\citep{wei2021finetuned}, T0~\citep{sanh2021multitask}, and SuperNI~\citep{wang2022super}, which assemble large instruction-tuning datasets from existing human-labeled datasets; and (2) synthesized data, like Self-Instruct~\citep{wang2023self}, WizardLM~\citep{xu2023wizardlm}, and GPT4-Alpaca~\citep{peng2023instruction}, which create instruction-tuning datasets by synthesizing from powerful LLMs like GPT-4~\citep{achiam2023gpt}. Both types of instruction-tuning data have advantages and disadvantages. Human-written data is limited in size due to the high cost and in task diversity because existing human-labeled datasets mostly focus on a few NLP tasks. Although synthesized data can be generated at any scale, the high rate of hallucination can lead to significant quality degradation. Moreover, the diversity of synthesized data is heavily influenced by the seed data. Without a diverse seed dataset, the synthesized data will lack domain coverage.

\paragraph{Mathematics Reasoning.} 
In recent years, there has been a growing interest in enhancing the mathematical reasoning abilities of large language models (LLMs). Three main approaches have been proposed to improve LLMs' mathematical reasoning skills:
\begin{itemize}[leftmargin=*]
\item Prompting: Chain-of-thought-prompting (CoT)~\citep{nye2022show,wei2022chain} elicits LLMs' inherent reasoning ability by demonstrating intermediate reasoning steps. Program-of-thoughts-prompting (PoT)~\citep{chen2023program,gao2023pal} utilizes tools to further augment LLMs' math reasoning abilities. Subsequent work~\citep{gou2023tora,toshniwal2024openmathinstruct,yuan2024advancing} combines CoT and PoT to maximize LLMs' reasoning ability.
\item Continued Training: Enabling LLMs to solve mathematical problems has been a long-standing challenge. MINERVA~\citep{lewkowycz2022solving} and Galactica~\citep{taylor2022galactica} were pioneers in continued training of LLMs to adapt to scientific domains for math and science reasoning. Open-source models like Llemma~\citep{azerbayev2023llemma}, DeepSeek-Math~\citep{shao2024deepseekmath}, and Intern-Math~\citep{ying2024internlm} have surpassed MINERVA and Galactica on math benchmarks. These approaches mainly rely on using an efficient classifier to recall documents from Common Crawl to retrieve a massive high-quality math-related corpus (>100B tokens) to enhance math reasoning.    
\item Instruction Tuning: Instruction tuning aims to enhance LLMs' math reasoning skills by efficiently training on human-annotated public datasets like GSM8K~\citep{cobbe2021training}, MATH~\citep{hendrycks2021measuring}, and MathQA~\citep{amini2019mathqa}. However, these datasets are often insufficient in size and diversity. Therefore, recent work~\citep{yu2023metamath,yue2023mammoth,toshniwal2024openmathinstruct,luo2023wizardmath,li2024common} proposes augmenting them with strong commercial LLMs like GPT-4~\citep{achiam2023gpt}. These methods can significantly boost LLMs' performance on in-domain math benchmarks but may fall short of generalization. 
\end{itemize}
Our work combines continued training with instruction tuning to exploit the benefits of both approaches. Specifically, our dataset is recalled from Common Crawl like DeepSeekMath. However, due to the significant level of noise in the raw corpus, we utilize a strong LLM to filter and clean the corpus to extract the instruction tuning pairs.   

\paragraph{Science Reasoning.}
In addition to mathematical reasoning, there is growing interest in improving LLMs' general scientific reasoning ability in subjects like physics, biology, chemistry, computer science, etc. Several benchmarks, such as MMLU~\citep{hendrycks2020measuring}, TheoremQA~\citep{chen2023theoremqa}, Sci-Bench~\citep{wang2023scibench}, SciEval~\citep{sun2023scieval}, and GPQA~\citep{rein2023gpqa}, have been developed to measure LLMs' reasoning ability on tasks beyond math. However, there has been less effort in curating high-quality training data for the science domain. Most datasets, like OpenWebMath~\citep{paster2023openwebmath}, Proof-Pile~\citep{azerbayev2023llemma}, and MathPile~\citep{zhang2024automathtext}, are heavily biased towards mathematics. In this work, we aim to generalize the pre-training data to broader subjects through our newly curated science seed data.